\documentclass[letterpaper, 10 pt, conference]{ieeeconf}  

\IEEEoverridecommandlockouts                              
\usepackage{graphicx} 

\title{\LARGE \bf
Fast and Robust Feature Matching for \mbox{RGB-D} Based Localization
}

\author{ Miguel Heredia$^{1}$, Felix Endres$^{2}$, Wolfram Burgard$^{2}$ and Rafael Sanz$^{1}$
\thanks{$^{1}$M. Heredia and R. Sanz are with the Department of Systems Engineering and Automation,
        University of Vigo, 36310 Vigo, Spain
        {\tt\small \{heredia, rsanz\} @ uvigo.es}}%
\thanks{$^{2}$F. Endres and W. Burgard are with the Department of Computer Science, University of Freiburg,
        Freiburg im Breisgau, 79110, Germany
        {\tt\small \{endres, burgard\} @ informatik.uni-freiburg.de}}%
\thanks{This work has partly been supported by the European Commission under the contract number FP7-ICT-248258-First-MM}
}

\usepackage{color}
\usepackage{hyperref}
\usepackage{url}

\begin{document}

\maketitle
\thispagestyle{empty}
\pagestyle{empty}

\begin{abstract}

In this paper we present a novel approach to global localization using an \mbox{RGB-D} camera in maps of visual features.
For large maps, the performance of pure image matching techniques decays in terms of robustness and computational cost.
Particularly, repeated occurrences of similar features due to repeating structure in the world (e.g., doorways, chairs, etc.) or missing associations between observations pose critical challenges to visual localization.
We address these challenges using a two-step approach. We first estimate a candidate pose using few correspondences between features of the current camera frame and the feature map.
The initial set of correspondences is established by proximity in feature space.
The initial pose estimate is used in the second step to guide spatial matching of  features in 3D, i.e., searching for associations where the image features are expected to be found in the map.
A RANSAC algorithm is used to compute a fine estimation of the pose from the correspondences.
Our approach clearly outperforms localization based on feature matching exclusively in feature space, both in terms of estimation accuracy and robustness to failure and allows for global localization in real time (30\,Hz).

\end{abstract}

\section{INTRODUCTION}

Localization is a fundamental requirement for most tasks in mobile robotics. The correct operation of mobile robots, such as UAVs, transportation robots or tour guides, is based on their ability to estimate their position and orientation in the world. 
SLAM approaches face the problem when no map is given and, therefore, the creation of the map and localization with respect to it have to be performed simultaneously. Given the map, fast and robust localization methods become necessary to efficiently perform localization within large environments.

\begin{figure}[t]
  \centering
  \includegraphics[width=\columnwidth]{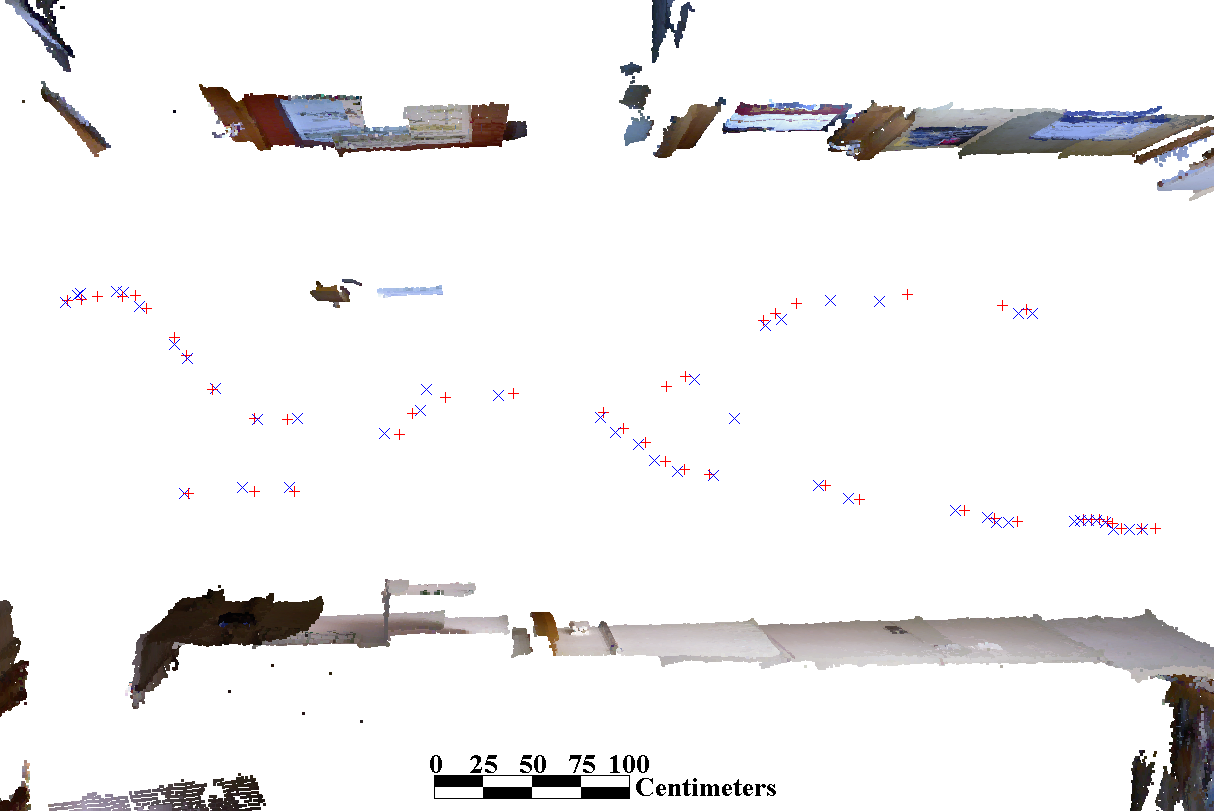}
  \caption{Our approach allows for highly precise global localization at frame rates up to 30\,Hz. Blue crosses represent the localization estimates, red pluses represent the ground truth.}
  \label{fig:pointcloud}
\end{figure}

The recent emergence of low cost versions of \mbox{RGB-D} sensors \cite{khoshelham12}, \cite{Langmann12}, has made them attractive as an alternative to the widely used laser scanners, in particular for inexpensive robotic applications.
Since such sensors gather dense 3D information of the scene at high frame rates (30Hz), it is possible to use them for six degrees of freedom (DoF) localization, which becomes necessary for robots whose 3D movement does not obey clear constrains that can be modeled beforehand. Recent approaches that perform \mbox{RGB-D} SLAM \cite{endres12icra} can create a database of visual features (Figure~\ref{fig:pointcloud}) as a sparse representation of the environment. This sparse map of features can be used to perform global six-DoF global localization using only a \mbox{RGB-D} sensor.

State-of-the-art methods to perform localization found in the literature are probabilistic filter algorithms, e.g., the extended Kalman filter (EKF) \cite{Cheeseman87} and particle filters~\cite{Fox99MCL}. Both use a Bayesian model to integrate sensor measurements (e.g., odometry, laser scans) into the current state representation. Particle filters overcome the severe scale limitations of Kalman filtering and can deal with highly non-linear models and non-Gaussian posteriors \cite{Thrun02particlefilters}. Particle filters are typically used in 2D and, therefore, particles need to represent a state with three DoF. In 3D localization, however, particles would have to be generated in a six DoF space, which dramatically increases the number of required particles. Most of approaches to visual localization are using feature descriptors-based feature matching to take perform data association by nearest neighbor searches in descriptor space. While this technique demonstrated to be useful and robust for matching two images to each other, it is prone to failure when matching against a huge database of features such as a large-scale map.

The contribution of this paper is a novel approach for vision-based global localization that performs independently from the previous state of the system using exclusively an \mbox{RGB-D} sensor and a sparse map of visual features.
Our system overcomes issues of traditional feature matching techniques when using large feature maps, such as the existence of different points of interest with similar descriptor values or the existence of the same keypoint in different objects of the same class (e.g., several chairs of the same model). 
In order to cope with this, we develop a two-step algorithm. In the first step we perform a fast guess of the sensor pose. We use it in the second step to allow for a more efficient spatial search for matching features. Using a spatial search method is possible due to the availability of depth information for the features. To ensure robust matching, we exploit information about the descriptor distances of matched features encountered during the map creation.
The results we obtain in terms of accuracy, robustness and execution time show that our approach is suitable to be used for online navigation of a mobile robot.

\begin{figure}[t]
  \centering
  \includegraphics[width=\columnwidth]{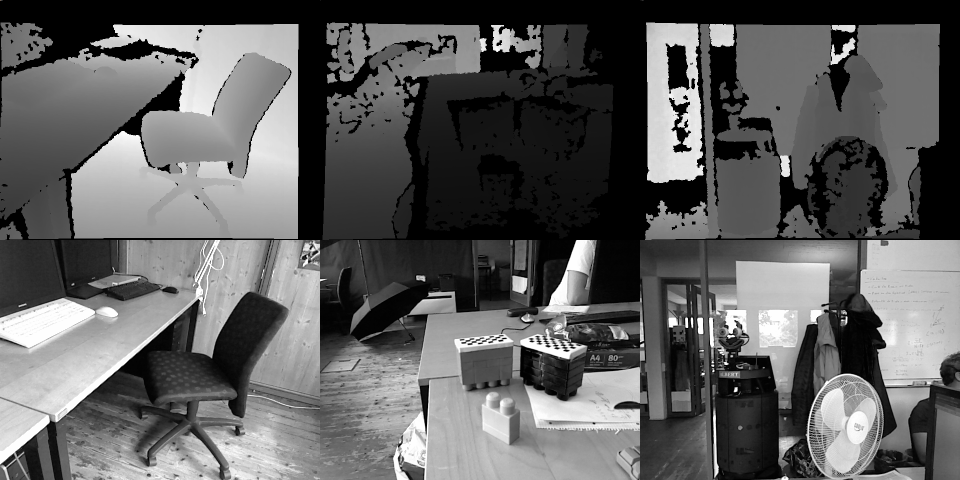}
  \caption{Input data: Depth and correspondent monochromatic images of several scenes of the laboratory, gathered with a Kinect sensor.}
  \label{fig:kinect_data}
\end{figure}

\section{RELATED WORK}

Finding the transformation given by a set of point correspondences is a common problem in computer vision, e.g., for ego-motion estimation. 
A method to get a closed-form solution by means of a Least-Squares Estimation (LSE) is given in~\cite{Umeyama91}. However, when dealing with sets of point correspondences containing wrong associations, the result given by a LSE is distorted by the outliers. This is problem is commonly addressed by using a sample consensus method such as RANSAC~\cite{fischler81}.

Zhang \emph{et al.} \cite{Zhang05CRV} provide an uncertainty estimation of a 3D stereo-based localization approach using a correspondence-based method to estimate the robot pose.
As in our work, visual features are extracted from the image and RANSAC is used to remove the outliers from the initial matches between features in two consecutive images.
In contrast, our approach establishes correspondences by image-to-map matching. 
Thus, additional sources of false correspondences arise, such as repeated objects in the world, the presence of several features in the map extracted from the same point in the world, or the much larger number of features in the map, which increases the chance for random matches.

Common methods to keep track of the camera pose found in the literature are visual odometry and SLAM techniques.
The term visual odometry was used for the first time in \cite{Nister04CVPR} to designate incremental motion estimation from visual information.
They propose two algorithms, for monocular and stereo visual odometry, and demonstrate their approach using a hand-held camera, an aerial vehicle and a ground vehicle.
They implement a kind of spatial filtering by performing bidirectional feature matching. We also take profit of a spatial-guided matching to reduce the possibilities of false matches and increase system speed.
Davison and Murray \cite{Murray98ECCV, Davison02PAMI} developed a 3D visual SLAM approach using active vision, based on the idea of a full-covariance Kalman Filter-based method presented in \cite{Davison98}.
They point out that the EKF cannot represent the multi-modal probability density functions resulting from errors in data association (mismatches).
Particle filters are a possible solution but they are considered infeasible due to the large increase in computational complexity with the dimension of the state vector.
They deal with misassociations by introducing a robust single feature tracking method. 
A consistency check is applied between the feature locations in the two images of the stereo system.
As well as in our approach, the goal of the feature map is the localization,
not the map itself, but the size of the map is bounded avoiding the problems
with large-scale maps.
Our approach establishes multiple matches, rejects false data associations and
can tolerate occurring mismatches.  We perform a further consistency check of
the relative locations between features to ensure that a small group of
features detected on the image was matched correctly against the feature map.

MCL overcomes the limitations of EKFs as mentioned earlier. It was successfully used in \cite{Fox99MCL} for vision-based localization of a tour-guide robot in a museum using a map of the ceiling and a camera pointing to it. In contrast to this approach, we do not rely on odometry measurements to predict the pose, and are not restricted to planar motion. Additionally, MCL would report incorrect locations after unexpected robot motions or sensor outages. Sensor Resetting Localization \cite{lenser00MCL} partially substitutes particles by new ones directly generated from the sensor measurements when the position estimate is uncertain. Mixture-MCL \cite{Thrun01Robust} combines standard MCL with dual-MCL to drastically reduce computational cost and localization error. Dual-MCL also generates particles from the current sensor measurements and was shown to deal well with the kidnapped robot problem when properly combined with standard MCL. We do not need a reset process, since our estimation is independent from the prior state.
Our approach could be used in combination with Monte Carlo Localization (MCL) for efficient particle initialization and weighting. 
However, the results of our experiments show that our estimate is accurate enough to be used as final result, without any filtering.


Howard~\cite{Howard08IROS} published an influential work in the field of real-time stereo visual odometry.  
His algorithm estimates the motion between consecutive stereo pairs.
Similarly to Nister \emph{et al.}~\cite{Nister04CVPR}, he performs a robust bidirectional matching.
However, the major contribution is the use of an alternative algorithm for inlier detection, rather than outlier rejection.
This algorithm is based on mutual consistency of the matches, similar to the one we perform when computing the coarse pose estimate.
His idea is to find the largest set of mutually consistent matches, using a consistency matrix.
Since this is a NP complete problem, a heuristic 
is used to reduce the computational complexity to $O(n^2)$.
According to the author, this method can easily cope with datasets containing 90\% outliers.

To the best of our knowledge, at the time being, there is no other dedicated global localization approach for the recently introduced \mbox{RGB-D} sensors.
However, a number of novel approaches for visual odometry have been proposed, which exploit the available combination of color, density of depth and the high frame rate to improve alignment performance as compared, e.g., to the iterative closest point algorithm~\cite{besl92icp}.
In \cite{dryanovski12icra} and \cite{pomerleau11tracking}~adaptations of ICP are proposed to process the high amounts of data more efficiently.  
Steinbruecker \emph{et al.}~\cite{steinbruecker11iccv} present a transformation estimation based on the minimization of an energy function.  For frames close to each other, they achieve enhanced runtime performance and accuracy compared to Generalized ICP~\cite{segal09rss}.
Using the distribution of normals, Osteen \emph{et al.}~\cite{osteen12icra} improve the initialization of ICP by efficiently computing the difference in orientation between two frames, which allows a substantial drift reduction.
These approaches work well for computing the transformation for small incremental changes between consecutive frames, but they are of limited applicability for global localization in a map.

Two works that use \mbox{RGB-D} sensors for 6 DoF SLAM are \cite{henry2012IJPR} and \cite{endres12icra}.
Both first perform a feature extraction and matching process to find correspondences between the current image and the previous one. 
RANSAC is used to robustly determine the motion between point clouds from feature matches.
In \cite{henry2012IJPR}, additionally, an ICP algorithm is performed to refine the transformation.  In both approaches a pose graph is created from the frame-to-frame transformation estimates and a global optimization of the map is carried out applying the toro~\cite{grisetti07iros} and the g2o~\cite{kuemmerle11icra} graph optimization frameworks, respectively.
In \cite{endres12icra}, the trajectory estimation is combined with the sensor data to create a 3D occupancy grid map of the environment, using the OctoMap framework \cite{hornung13auro}. 
In our approach, we use a map of landmarks, i.e. 64D visual feature descriptors with 3D positions. Further we rely on information about the distance between the descriptors, when they were matched during mapping.


\section{METHODOLOGY}

\begin{figure}[t]
  \centering
  \includegraphics[width=\columnwidth]{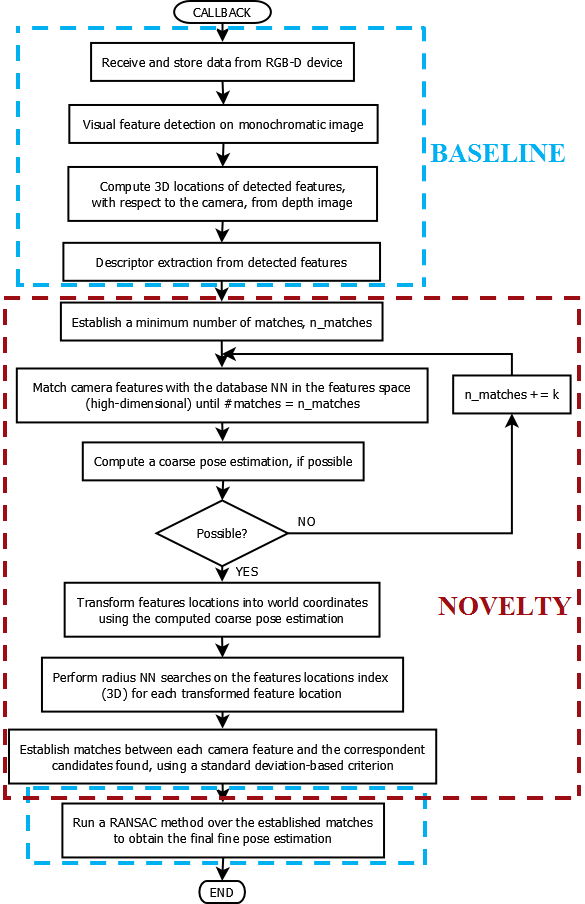}
  \caption{
Block diagram of our approach. First tasks are shared with existing image matching approaches: Detection of visual features and descriptor extraction. The later process of data association is the main novelty of this work: First a certain number of correspondences are established in descriptor space and used to compute a coarse pose estimate. The coarse pose estimate allows for nearest neighbor searches in 3D space. Candidates found are considered according to the proximity between correspondent features in descriptor space. These correspondences are used to get an accurate pose estimate by means of a RANSAC method.
}
  \label{fig:diagram}
\end{figure}
Our goal is to localize a \mbox{RGB-D} sensor in a given map of the world using only the data provided by the device.
Our approach (schematically depicted in Figure~\ref{fig:diagram}) performs the following high-level tasks: First visual features are detected in the monochromatic image.
Then, a descriptor is computed for each feature.
These descriptors are used to perform data association between the features detected in the image and those contained in a feature map of the world.
This matching process is based on proximity between features in the high-dimensional feature space.
Correspondences are used to compute a coarse estimation of the sensor pose.
This estimation is used to guide nearest neighbor searches in the 3-D space looking for each feature found in the image where it is expected to be in the map.
Matches established this way are used to compute a fine pose estimation.

Nearest neighbor searches are an important task in our approach.
Searches are performed both in the high dimensional feature space and in 3D space.
During the development of our system they showed to be the most time-consuming task.
The OpenCV~\cite{opencv_library} implementation of the FLANN library provided by Muja and Lowe \cite{Muja09FLANN} is used to perform approximate nearest neighbor (ANN) searches.
The search performance is highly dependent on the used data structure and algorithm.
A series of experiments were performed to optimize the search for use with large maps of up to $150,000$ features.
This optimization process successfully reduced the runtime for nearest neighbor search up to 25\% in the 3D index and up to 9\% in the high-dimensionality descriptor space.
However, the ANN search remains the computational bottleneck and reducing it is the main focus of the presented approach.


The initial steps of our approach are similar to an image matching process.
FLANN indices are created offline, in order to minimize searching time during operation.
The main differences lie in the process of pose estimation from the sensor data, which is the major contribution of this work (Figure~\ref{fig:diagram}).
Figure~\ref{fig:kinect_data} shows some pairs of typical data from the Kinect gathered during experiments.
The first step is the visual feature detection on the monochromatic image using SURF \cite{Bay08}.
Then, SURF descriptors are computed for each one of the detected features.
The next step is a coarse pose estimation process.
To obtain a coarse estimation of the camera pose with respect to the world, it is necessary to establish several initial matches.
Therefore, the process presents two steps that are repeated until a good coarse estimation is found.
First some matches between features detected on the camera image and map features are established.
The correspondent feature in the map is the nearest neighbor on the 64-dimensional feature space (the SURF features used in our implementation have 64 dimensions).
Then, this small group of matches is used to compute the coarse pose estimation.
This process involves finding the largest subset of mutually consistent matches in a similar way to \cite{Howard08IROS}.
If the matches are not good enough to produce a satisfactory coarse estimation, more features from the camera image are matched against the feature database and the estimation is computed again.
The process ends when an acceptable estimation is reached or when all the features detected on the image have been already matched.
Once the coarse pose of the sensor is estimated, it is used to predict where the features observed by the sensor are with respect to the map frame.
For each feature, all the map features located in a certain neighborhood of the predicted spatial location are taken as match candidates.
This matching process guided by spatial information is faster than descriptor-based matching, because the dimensionality is much lower. 

Given a sensor reading with $N$ features, a standard matching approach would take $ N \times t_\mathrm{64D}$, where $t_\mathrm{64D}$ is the search time in the descriptor space.
In contrast, the overall runtime for matching features in our approach is 
$$ M \times t_\mathrm{64D} + N \times t_\mathrm{3D}.$$
Where $M$ is the average number of nearest neighbors required to establish the coarse pose estimation and $t_\mathrm{3D}$ is the spatial nearest neighbor search time per feature.
In an analysis on selecting the optimal approximate nearest neighbor algorithm for matching SIFT features to a large database, 
Muja and Lowe~\cite{Muja09FLANN} achieved an speedup factor of 30 with respect to a linear search, for 90\,\% search precision. 
Since $$M \ll N \quad\mathrm{and}\quad t_\mathrm{3D} < t_\mathrm{64D}$$ our approach outperforms a standard matching approach (timings for $t_\mathrm{3D}$ and $t_\mathrm{64D}$ are given in Table~\ref{tab:runtimes}).
For each candidate feature found, a match between the camera feature and this candidate is established only if the Euclidean distance between them in descriptor space is lower than a threshold.
This threshold is set to the standard deviation observed when matching this feature during the mapping process.
Finally, a RANSAC method is used to compute the final estimation of the camera pose with respect to the map.


\section{EXPERIMENTS}

\begin{figure*}[t]
      \centering
	\includegraphics[width=\textwidth]{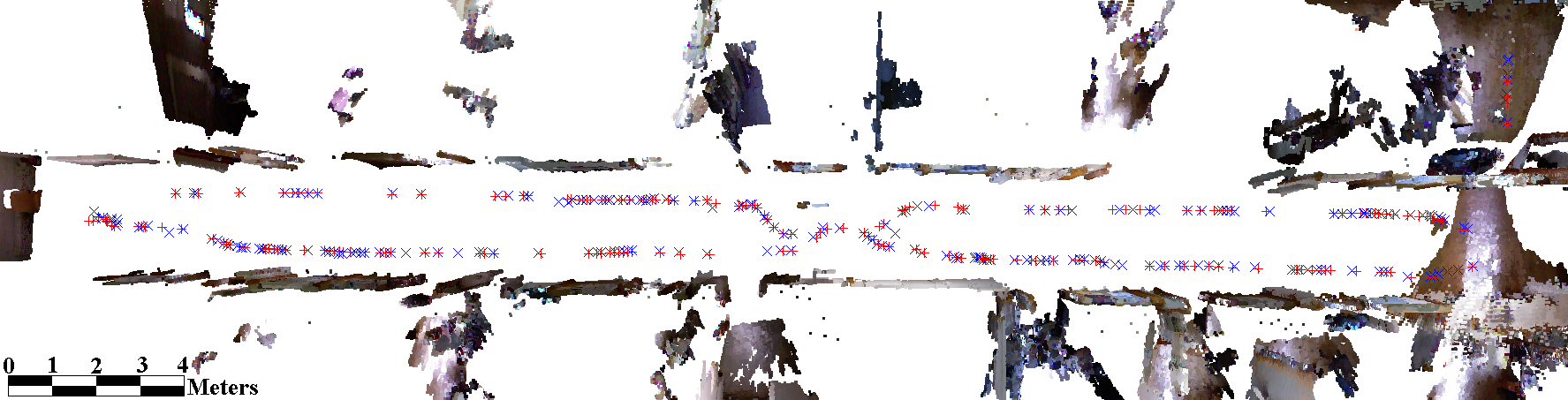}
      	\caption{Trajectory of the robot, top view. Blue crosses represent the localization estimates, red pluses represent the ground truth.}
      	\label{fig:trajectories}
\end{figure*}

In this section we evaluate our approach in terms of accuracy of the estimate with respect to the ground truth, robustness of the estimation method and process runtimes. 
We present experimental results comparing the proposed approach to a baseline method of performing feature matching based exclusively on proximity on feature space.

This baseline approach represents a standard way to face the problem of visual localization, as in a image matching process.
As illustrated in Figure~\ref{fig:diagram}, it consists of the following basic phases: Keypoint detection and descriptor extraction, for which we use the SURF implementation from OpenCV.
Then, the detected features are matched against the nearest map features in the 64D space using the OpenCV FLANN library.
Finally, the camera pose with respect to the map is computed using RANSAC for robustness. 

The goal of our approach is to estimate the Kinect sensor pose.
In order to evaluate the quality of the estimation, the result is compared to a ground truth sensor pose for each estimate.
In order to get the ground truth trajectory, we use a PR2 robot with a Kinect sensor mounted on its head. 
The PR2 base laser scanner is used to obtain a highly accurate 2D position estimate for the robot. 
To record the data for our experiments, the PR2 was navigated through the laboratory of the AIS department along the corridor of the ground floor.
A top view of the environment is shown in Figure~\ref{fig:trajectories}.

To evaluate our approach, we gathered two datasets of the same area, which we use for \emph{training} and \emph{evaluation}, i.e., one to create the map and the other one to evaluate our algorithm.
The robot pose estimate obtained from the laser range measurements is used to create the feature map from the training data and as ground truth for the evaluation dataset. 
Nevertheless, the feature matching process is still performed during mapping to obtain standard deviations for matching features. 
The resulting map is stored as a database of feature locations, feature descriptors and the descriptor matching deviation.

The number of nearest neighbors considered for correspondences determines the tradeoff between the success rate and the runtime.
Therefore three different versions of the baseline approach are evaluated, performing single nearest neighbor matching, ten nearest neighbor matching and 20 nearest neighbor matching, respectively.

\begin{figure}[t]
  \centering
  \includegraphics[width=\columnwidth]{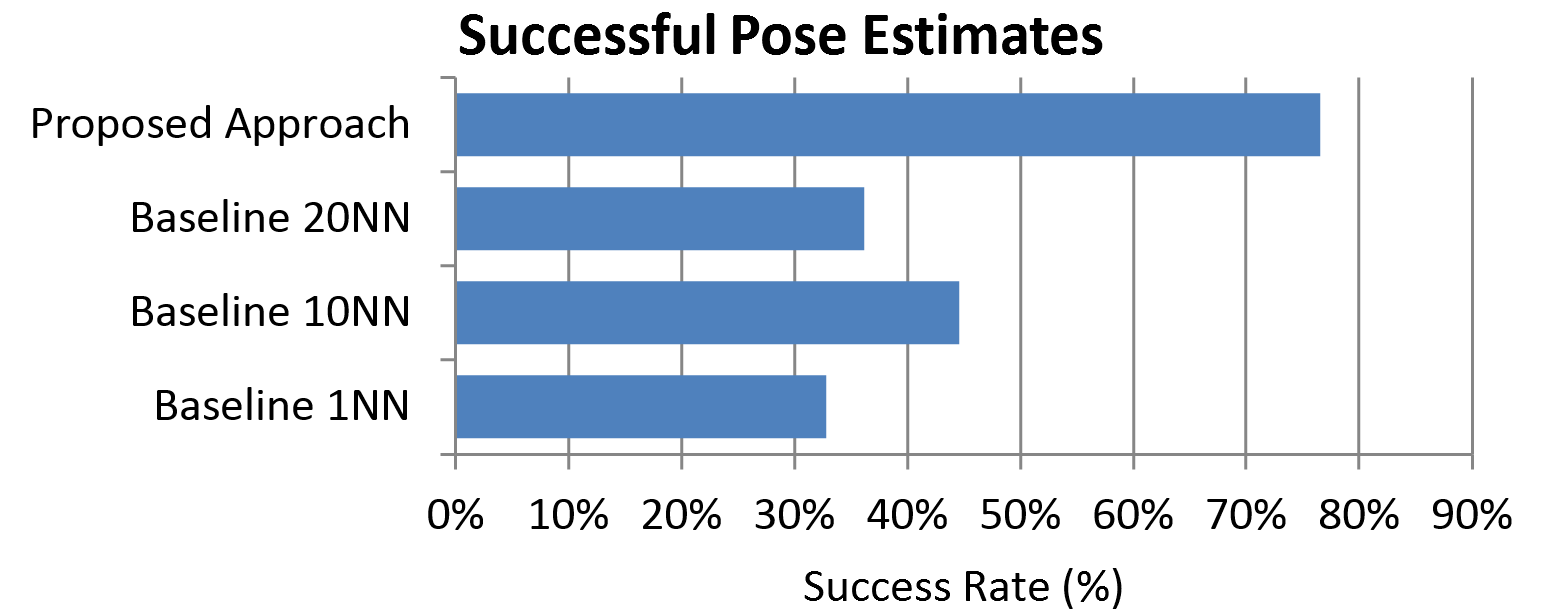}
  \caption{Robustness comparison: Percentage of successful estimates per approach. Note the low robustness of the baseline approaches}
  \label{fig:robustness}
\end{figure}

Robustness is given by the number of successful final estimates of the camera pose with respect to the total number of tries. 
An estimation is considered as failed if the translational error along any direction is higher than 0.5\,m or no pose can be computed.
Failures are not taken into account to calculate the RMSE values.
Figure~\ref{fig:robustness} shows the robustness values for the three versions of the baseline approach and the proposed approach as percentage of success.
Notice how the proposed approach widely outperforms the baseline approaches.

\subsection{Accuracy}

\begin{figure}[t]
  \centering
  \includegraphics[width=\columnwidth]{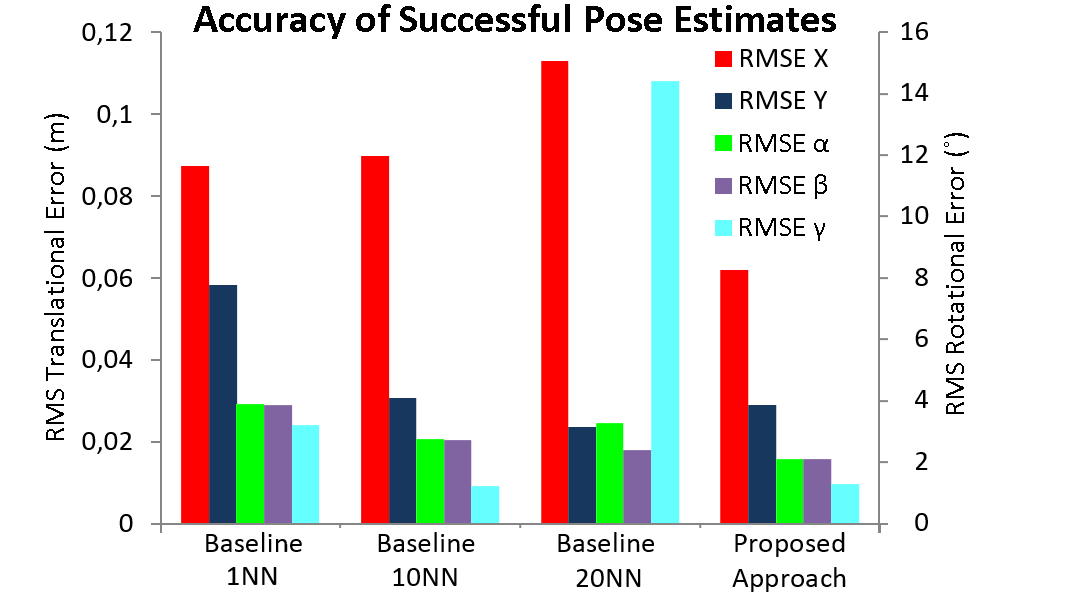}
  \caption{Error comparison: Translational and rotational RMS errors of each approach. Note the high values of translational RMSE in the main direction of the motion ($X$) for the baseline approaches}
  \label{fig:error}
\end{figure}

Accuracy is evaluated by computing the root mean square errors (RMSE) of the pose estimate (blue crosses in Figure~\ref{fig:trajectories}) with respect to the ground truth pose (red pluses in Figure~\ref{fig:trajectories}).
Let $XYZ$ be the fixed coordinate frame of the map, where the $X$ axis is parallel to the main direction of the motion --i.e., the long corridor described above-- and the $Y$ axis is perpendicular to $X$, both in the horizontal plane. Let $UVW$ be the mobile frame that represents the robot pose, estimated by means of the laser range scanner (ground truth), and $U'V'W'$ the same frame estimated using the considered approach. The translational errors “RMSE $X$” and “RMSE $Y$” refer to the RMS value of the distance between the origins of $UVW$ and $U'V'W'$ along the respective axis of the map frame. The rotational errors “RMSE $\alpha$”, “RMSE $\beta$” and “RMSE $\gamma$” are the RMS values of the angle between the axis $U$ and $U'$, $V$ and $V'$, $W$ and $W'$, respectively.
Figure~\ref{fig:error} shows the accuracy results obtained for each approach. 

\subsection{Runtime}
Table~\ref{tab:runtimes} shows all the runtimes of the relevant components of our approach and compares them to the equivalent runtimes registered for the baseline approaches.
It is easy to observe that our dual nearest neighbor search is faster than a single search in descriptor space, due to the low number of correspondences we need to compute a coarse pose estimate.

\begin{table}[t]
\caption{Runtime results of the experiment, specified by main tasks of the approaches. The performance of our approach is compared to the performance of the baseline approaches}
\label{tab:runtimes}
\begin{center}
\begin{tabular}{l c c c c}
\hline
  &\multicolumn{4}{c}{Average Runtimes (s)}\\
\hline
Program Tasks&Base. 1&Base.10&Base. 20&Proposal\\
\hline
NN Search in 64D&0.0233&0.0242&0.0219&0.0014 \\ 
Coarse Pose Estimation&---&---&---&0.0005 \\ 
Radius NN Search in 3D&---&---&---&0.0155\\
RANSAC Estimation&0.0092&2.26&7.06&0.0149\\
Total Runtime&0.0325&2.28&7.08&0.0323\\
\hline
\end{tabular}
\end{center}
\end{table}

The approach we introduce in this work outperforms all the versions of the baseline in terms of accuracy and robustness. The translational RMS error along the $X$ axis is 6.2\,cm and only 2.9\,cm along the $Y$ axis. Note that the translational RMSE $Y$ is lower for the third baseline experiment, but the rest of the measurements show higher errors than in our approach. Apart from that, the poor robustness of this baseline version makes our approach the best alternative. All baseline approaches exhibit values of the ratio below 50\%. In contrast, our approach produces a successful estimate in 77\% of the cases. Further, the rest of the cases do not lead to a wrong estimate. In most cases, no coarse pose estimate can be computed due to a lack of mutual coherence between matches. Therefore, the rest of the pose estimation process can be skipped, leading to fast runtimes for failure cases. In terms of runtime, our algorithm shows an average runtime of 32.3\,ms. This means that a rate of 30\,Hz (sensor rate) is reachable if the feature detection and the descriptors extraction run in parallel (e.g., in a GPU) to the process.

\section{CONCLUSIONS}

In this paper we presented an approach to estimate the pose of a robot using a single \mbox{RGB-D} sensor. Visual features are extracted from the monochromatic images. Nearest neighbor searches in the feature descriptor space are carried out to generate sets of correspondences between the image and the feature map. A restrictive coherency check based on preservation of mutual distances between features in the 3D space is carried out to reach a small set of high quality correspondences. These correspondences are used to compute a coarse sensor pose estimate, which guides a spatial matching process. The final pose estimation is computed from the correspondences obtained from the spatial-guided matching.

We carried out experiments for global evaluation of the approach using real data gathered from an office environment, large enough to generate a high number of features in the feature map. Our approach showed to deal well with large feature maps, even when several features are present in the map to represent the same point in the world. It also works properly through environments with repeated objects –chairs, tables, screens, etc. – and repetitive structure. It outperforms methods based on feature matching only in feature space, taken as baseline approaches. In our experiments, the number of successful pose estimates is 54\,\% higher than the number of correct estimations provided by the best baseline approach for the same dataset. The maximum translational RMS error is 6.2\,cm and the maximum rotational RMS error is 2.1$^\circ$. In the current implementation the runtime was 0.5\,seconds, but this value is largely influenced by the processes of features detection and descriptors extraction. If these processes are performed fast enough in parallel –for example in a GPU– with the estimation process, our approach is able to operate in real time (30Hz). In conclusion, the approach we introduced here is better in terms of accuracy, robustness and runtime than approaches that perform features matching based only in proximity between descriptors.

\section{ACKNOWLEDGMENT}

We would like to thank the people of the University of Freiburg (Germany) for their assistance during Miguel Heredia’s visit to the Autonomous Intelligent Systems Laboratory.


\bibliographystyle{IEEEtran}
\bibliography{references}

\end{document}